\documentclass[final]{style/cvpr}

\usepackage{times}
\usepackage{epsfig}
\usepackage{graphicx}
\usepackage{amsmath}
\usepackage{amssymb}
\usepackage{mwe}

\usepackage{enumitem}
\usepackage{tabularx}
\usepackage{graphicx}
\usepackage{booktabs}
\usepackage{graphbox}
\usepackage[font=small]{caption}

\usepackage{xpatch}
\makeatletter
\xpatchcmd{\paragraph}{3.25ex \@plus1ex \@minus.2ex}{3pt plus 5pt minus 1pt}{\typeout{success!}}{\typeout{failure!}}
\makeatother

\usepackage[pagebackref=true,breaklinks=true,colorlinks,bookmarks=false]{hyperref}

\begin{document}

\title{Animating Pictures with Eulerian Motion Fields}

\newlength{\spacebetweenauthors}
\setlength{\spacebetweenauthors}{13mm}
\author{Aleksander Holynski$^1$  \hspace{\spacebetweenauthors} Brian Curless$^1$ \hspace{\spacebetweenauthors} Steven M. Seitz$^1$  \hspace{\spacebetweenauthors} Richard Szeliski$^2$  \vspace{1em}\\ 

$^1$University of Washington \hspace{\spacebetweenauthors} $^2$Facebook  \\
\href{eulerian.cs.washington.edu}{\texttt{eulerian.cs.washington.edu}}\vspace{-0.5em}

} 

\twocolumn[{
\renewcommand\twocolumn[1][]{#1}
\maketitle
\centering
    \setlength\fboxsep{0pt}
\includegraphics[align=c,width=\linewidth]{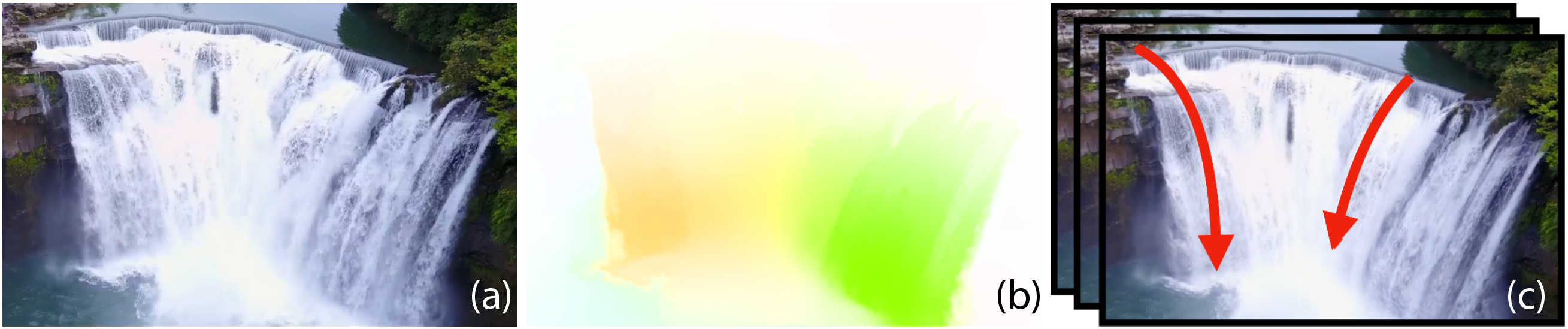}
\captionof{figure}{Given a single input image (a), our method estimates an image-aligned motion field (b), and uses it to create a looping video (c).}
\vspace{2mm}
  \label{fig:teaser}

}]

\begin{abstract}
\vspace{-2mm}
In this paper, we demonstrate a fully automatic method for converting a still image into a realistic animated looping video. We target scenes with continuous fluid motion, such as flowing water and billowing smoke. Our method relies on the observation that this type of natural motion can be convincingly reproduced from a static Eulerian motion description, i.e. a single, temporally constant flow field that defines the immediate motion of a particle at a given 2D location. We use an image-to-image translation network to encode motion priors of natural scenes collected from online videos, so that for a new photo, we can synthesize a corresponding motion field. The image is then animated using the generated motion through a deep warping technique: pixels are encoded as deep features, those features are warped via Eulerian motion, and the resulting warped feature maps are decoded as images. In order to produce continuous, seamlessly looping video textures, we propose a novel video looping technique that flows features both forward and backward in time and then blends the results. We demonstrate the effectiveness and robustness of our method by applying it to a large collection of examples including beaches, waterfalls, and flowing rivers.
\end{abstract}
\vspace{-5mm}

\section{Introduction}

For humans, a picture often contains much more than a collection of pixels. Drawing from our previous observations of the world, we can recognize objects, structure, and even imagine how the scene was moving when the picture was taken. Using these priors, we can often envision the image as if it were animated, with smoke billowing out of a chimney, or waves rippling across a lake. In this paper, we propose a system that learns these same motion priors from videos of real scenes, enabling the synthesis of plausible motions for a novel static image and allowing us to render an animated video of the scene. 

General scene motion is highly complex, involving perspective effects, occlusions, and transience. For the purposes of this paper, we restrict our attention to 
fluid motions, such as smoke, water, and clouds, which are well approximated by Eulerian motion, in particular, particle motion through a static velocity field.

Our proposed method takes as input a single static image and produces a looping video texture. We begin by using an image-to-image translation network \cite{wang2018high} to synthesize an Eulerian motion field. This network is trained using pairs of images and motion fields, which are extracted from a large collection of online stock footage videos of natural scenes. Through Euler integration, this motion field defines each source pixel's trajectory through the output video sequence. 
Given the source pixel positions in a future frame, we render the corresponding frame using a deep warping technique: we use an encoder network to transform the input image into a deep feature map, warp those features using a novel \emph{temporally symmetric} splatting technique, and use a decoder network to recover the corresponding warped color image. Lastly, in order to ensure our output video loops seamlessly, we apply a novel video looping technique that operates in deep feature space.

Our contributions include (1) a novel motion representation for single-frame textural animation that uses Euler integration to simulate motion, (2) a novel \emph{symmetric} deep splatting technique for synthesizing realistic warped frames, and (3) a novel technique for seamless video looping of textural motion.

\section{Previous Work}
\label{sec:prior_work}
\subsection{Video Textures}
There is a large body of work aimed at producing 
looping videos, known variously as {\em video textures}, {\em cinemagraphs}, or {\em live photos}. 
These techniques typically take as input a longer video sequence and, through some analysis of the motion, 
produce a single seamlessly looping video, or an infinite (yet not obviously looping) video
\cite{schodl2000video}. The term {\em cinemagraph} often refers to selective animation of looping clips, where only certain parts of the frame, as chosen by the user, are animated (or de-animated) \cite{bai2012selectively}. Newer approaches \cite{tompkin2011towards,yeh2012approach,liao2013automated,liao2015fast,oh2017personalized} perform this task fully automatically, determining which regions are easily looped, and which regions contain motions that are large in magnitude or otherwise unsuitable for looping. These approaches have also been extended to operate on specific domains, such as videos of faces \cite{bai2012selectively}, urban environments \cite{yan2017turning}, panoramas \cite{agarwala2005panoramic}, and continuous particle effects \cite{bhat2014flow, lin2019creating}. All these methods, however, 
require a video as input.

\subsection{Single-image animation}
There are also a number of methods aimed at animating still images.  Recently, these techniques have gained popularity through commercial applications such as Plotagraph\footnote{https://app.plotaverse.com} and Pixaloop\footnote{https://www.pixaloopapp.com}, which allow users to manually ``paint'' motion onto an image. 
In the following, we focus on approaches to perform some of this annotation automatically. 

\paragraph{Physical simulation.} Instead of manually annotating the direction and magnitude of motion, the motion of certain objects, such as boats rocking on the water, can be physically simulated, as long as each object's identity is known and its extent is precisely defined \cite{chuang2005animating}, or automatically identified through class-specific heuristics \cite{jhou2015animating}. Since each object category is modeled independently, these methods do not easily extend to more general scene animation.

\paragraph{Using videos as guidance.}
Alternatively, motion or appearance information can be transferred from a user-provided reference video, containing either similar scene composition \cite{prashnani2017phase}, aligned information from a different domain, such as semantic labels \cite{wang2019few}, or unaligned samples from the same domain \cite{siarohin2019first, cheng2020time}.
Instead of a single user-provided video, a database of homogeneous videos can be used to inherit nearest-neighbor textural motion, assuming a segmentation of the dynamic region is provided \cite{okabe2009animating}. 

\paragraph{Transformations in latent space.}
Recent advances in deep learning have enabled realistic, high-resolution image synthesis using generative adverserial networks (GANs). Many of these systems operate by representing images or scenes as a latent feature vector, which is decoded into a synthesized image. By perturbing the latent vector, or performing a randomized walk in the latent feature space, the resulting decoded images remain plausible, while also varying temporally \cite{shaham2019singan, hinz2020improved, karras2020analyzing}. These animations can visualize the space of possible appearances, but do not necessarily animate plausible motion. 

Instead of a random walk, one can also directly control movement by applying spatial warps to latent features \cite{logacheva2020deeplandscape}. Still, deciding \emph{how} to warp the image is non-trivial --- to produce a realistic video, the applied transformations must correspond with feasible motion in the scene. 

\paragraph{Using learned motion or appearance priors.}
Deep learning also enables motion synthesis from single-frame inputs \cite{gao2018im2flow, walker2015dense}. Similarly, video prediction methods \cite{zhang2020dtvnet, xue2016visual, li2018flow, xiong2018learning, pan2019video} can predict future video frames from a single image, even modelling the inherent multi-modality of predicting the future. These techniques typically predict a set of future frames at once, and thus are limited to either low spatial resolution or few predicted frames. 

Most similar to our work, Endo et al.~\cite{endo2019animatinglandscape} demonstrate high-quality motion and appearance synthesis for animating timelapses from static landscape imagery. 
In our evaluations, we provide comparisons to this technique, showing that our method more reliably estimates motion for scenes with fluids and animates videos with fewer visible artifacts.

\begin{figure*}
    \centering
    \includegraphics[width=\linewidth]{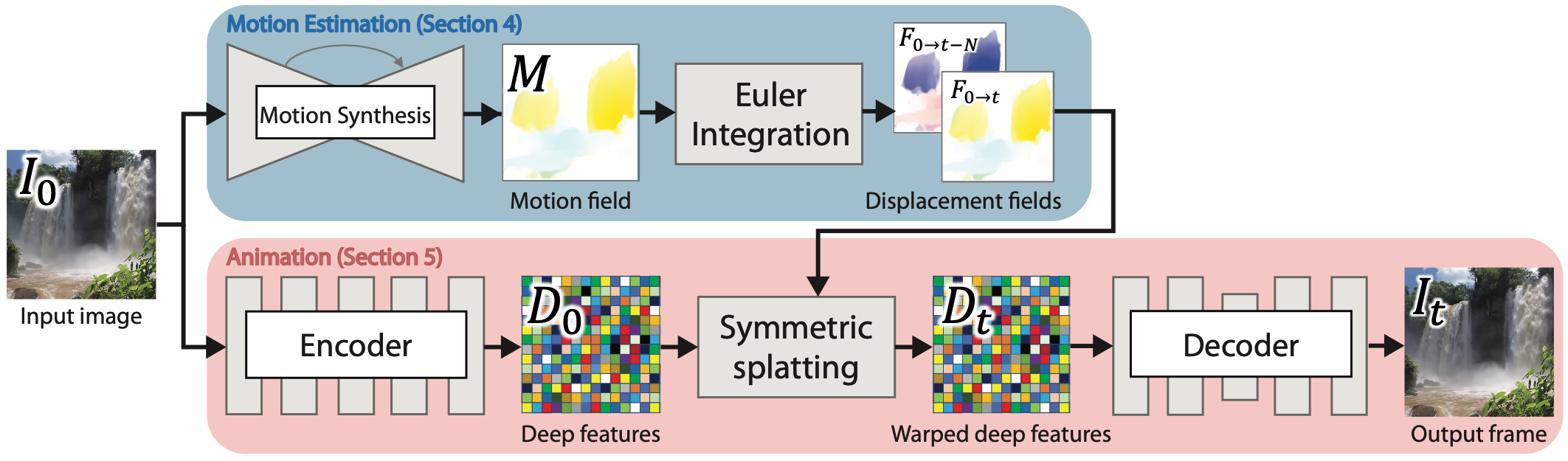}
    \caption{\small \textbf{Overview:} Given an input image $I_0$, our motion estimation network predicts a motion field $M$. Through Euler integration, $M$ is used to generate future and past displacement fields $F_{0\rightarrow t}$ and $F_{0\rightarrow t-N}$ , which define the source pixel locations in all other frames $t$. To animate the input image using our estimated motion, we first use a feature encoder network to encode the image as a feature map $D_0$. This feature map is warped by the displacement fields (using a novel symmetric splatting technique) to produce the corresponding warped feature map $D_t$. The warped features are provided to the decoder network to create the output video frame $I_t$. }    
    
    \label{fig:pipeline}
\end{figure*}

\section{Overview}
\label{sec:overview}

Given a single static image $I_0$, we generate a looping video of length $N+1$, consisting of frames $I_t$ with $t\in [0,N]$. Our pipeline begins by using an image-to-image translation network to estimate a corresponding motion field $M$ (Section~\ref{sec:motion_estimation}), which is used to define the position of each pixel in all future frames. We use this information to animate the image through a deep warping technique (Section~\ref{sec:animation}). Finally, in order to produce seamlessly looping videos, we introduce a technique to ensure that our videos always start and end with the same frame (Section~\ref{sec:looping}). Our approach is summarized in Figure~\ref{fig:pipeline}. 

\section{Motion estimation}
\label{sec:motion_estimation}
We begin by describing the motion model and the motion estimation network. Given an image as input, we wish to synthesize plausible motion for the observed scene. Prior work accomplishes this task through recurrent prediction of incremental flow fields \cite{endo2019animatinglandscape}, theoretically enabling generation of an infinite number of future frames at high-resolution. In practice, however, recurrent estimation often results in long-term distortion, since predicted motions are dependent on previously generated frames. In contrast, our motion field is only predicted once, given the input image, and thus does not degrade over time. Even though we use a single static motion field to represent the motion of an entire video, we can still model complex motion paths. This is because our motion field $M$ is a static \emph{Eulerian} flow field, i.e., a 2D map of motion vectors where each pixel's value defines its immediate velocity, which does not change over time. We use $M$ to simulate the motion of a point (particle) from one frame to the next via Euler integration:
\begin{equation} \label{eq:coordinate_eul}
    \hat{\mathbf{x}}_{t+1} = \hat{\mathbf{x}}_t + M(\hat{\mathbf{x}}_t) ,
\end{equation}
where $\hat{\mathbf{x}}_t$ is the point's $(x,y)$ coordinate in frame $t$. In other words, treating each pixel as a particle, this motion field is the flow between each frame and its adjacent future frame:
\begin{equation} \label{eq:flow_define}
    M(\hat{\mathbf{x}}_t) = F_{t\rightarrow t+1}(\hat{\mathbf{x}}_t)
\end{equation}

To synthesize this motion field, we train an image-to-image translation network \cite{wang2018high} on color-motion pairs, such that when provided with a new color image $I_0$, it estimates a plausible motion field $M$. Given an image, $M$ is only estimated once through an inference call to the network. Once estimated, it can be used to define the source pixel positions in all future frames $t$ by recursively applying:
\begin{equation} \label{eq:eulerian}
    F_{0\rightarrow t}(\hat{\mathbf{x}}_0) = F_{0\rightarrow t-1}(\hat{\mathbf{x}}_0) + M(\hat{\mathbf{x}}_0 + F_{0\rightarrow t-1}(\hat{\mathbf{x}}_0))
\end{equation}
This results in displacement fields $F_{0\rightarrow t}$, which define the trajectory of each source pixel in $I_0$ across future frames $I_t$. These displacement fields are then used for warping the input image, as further described in Section~\ref{sec:animation}. Computing $F_{0\rightarrow t}$ does not incur additional calls to the network --- it only uses information from the already-estimated $M$.

Note that unlike Endo et al. \cite{endo2019animatinglandscape}, who predict backward flow fields for warping (i.e., using bilinear backward sampling), we predict the \emph{forward} motion field, i.e., aligned with the input image. In our evaluations, we show that predicting forward motion results in more reliable motion prediction and sharper motion estimates at object boundaries. As a result, this enables more realistic animation of nearby scenes with partial occlusions, since regions that are moving are more precisely delineated from those that are not.

\section{Animation}
\label{sec:animation}
Once we have estimated the displacement fields $F_{0\rightarrow t}$ from the input image to all future frames, we use this information to animate the image. Typically, forward warping, i.e., warping an image with a pixel-aligned displacement field, is accomplished through a process known as splatting. This process involves sampling each pixel in the input image, computing its destination coordinate as its initial position plus displacement, and finally assigning the source pixel's value to the destination coordinate. Warping an image with splatting unfortunately suffers from two significant artifacts: (1) the output is seldom dense --- it usually contains holes, which are regions to which no source pixel is displaced, and (2) multiple source pixels may map to the same destination pixel, resulting in loss of detail or aliasing. Additionally, the predicted motion fields may be imperfect, and naively warping the input image can result in boundary artifacts. In the following section, we introduce a deep image warping approach to resolve these issues.

\subsection{Deep image warping}
\label{sec:warping}
Given an image $I_0$ and a displacement field $F_{0\rightarrow t}$, we adopt a deep warping technique to realistically warp the input frame and fill unknown regions. Our method consists of three steps: (1) use an encoder network to encode the input image $I_0$ as a deep feature map $D_0$, (2) use the estimated displacement field $F_{0\rightarrow t}$ to splat those features to a future frame, producing $D_t$, and (3) use a decoder network to convert the warped features to an output color image $I_t$. For our encoder and decoder networks, we use variants of the architectures proposed in SynSin \cite{wiles2020synsin}. More implementation details are provided in Section~\ref{sec:implementation_details}.

As mentioned in the previous section, unlike backward warping, splatting may result in multiple source pixels mapping to the same destination coordinate. In these cases, it is necessary to decide which value will occupy the pixel in the destination image. 
For this, we adopt softmax splatting \cite{niklaus2020softmax}, which assigns a per-pixel weighting metric $Z$ to the source image, and uses a softmax to determine the contributions of colliding source pixels in the destination frame:
\begin{equation}
\label{eq:splatting_metric}
D_t({\hat{\mathbf{x}}'}) = \frac{\sum_{\hat{\mathbf{x}}\in \mathcal{X}} D_0(\hat{\mathbf{x}}) \cdot \text{exp}(Z(\hat{\mathbf{x}}))}{\sum_{\hat{\mathbf{x}}\in \mathcal{X}} \text{exp}(Z(\hat{\mathbf{x}}))}
\end{equation}
where $\mathcal{X}$ is the set of pixels which map to destination pixel ${\hat{\mathbf{x}}'}$. 
Our method infers $Z$ automatically as an additional channel of the encoded feature map. The learned metric allows the network to assign importance to certain features over others, and the softmax exponentiation avoids uniform blending, resulting in sharper synthesized frames.

\paragraph{Symmetric Splatting.}
\begin{figure}
    \centering
    \includegraphics[width=\linewidth]{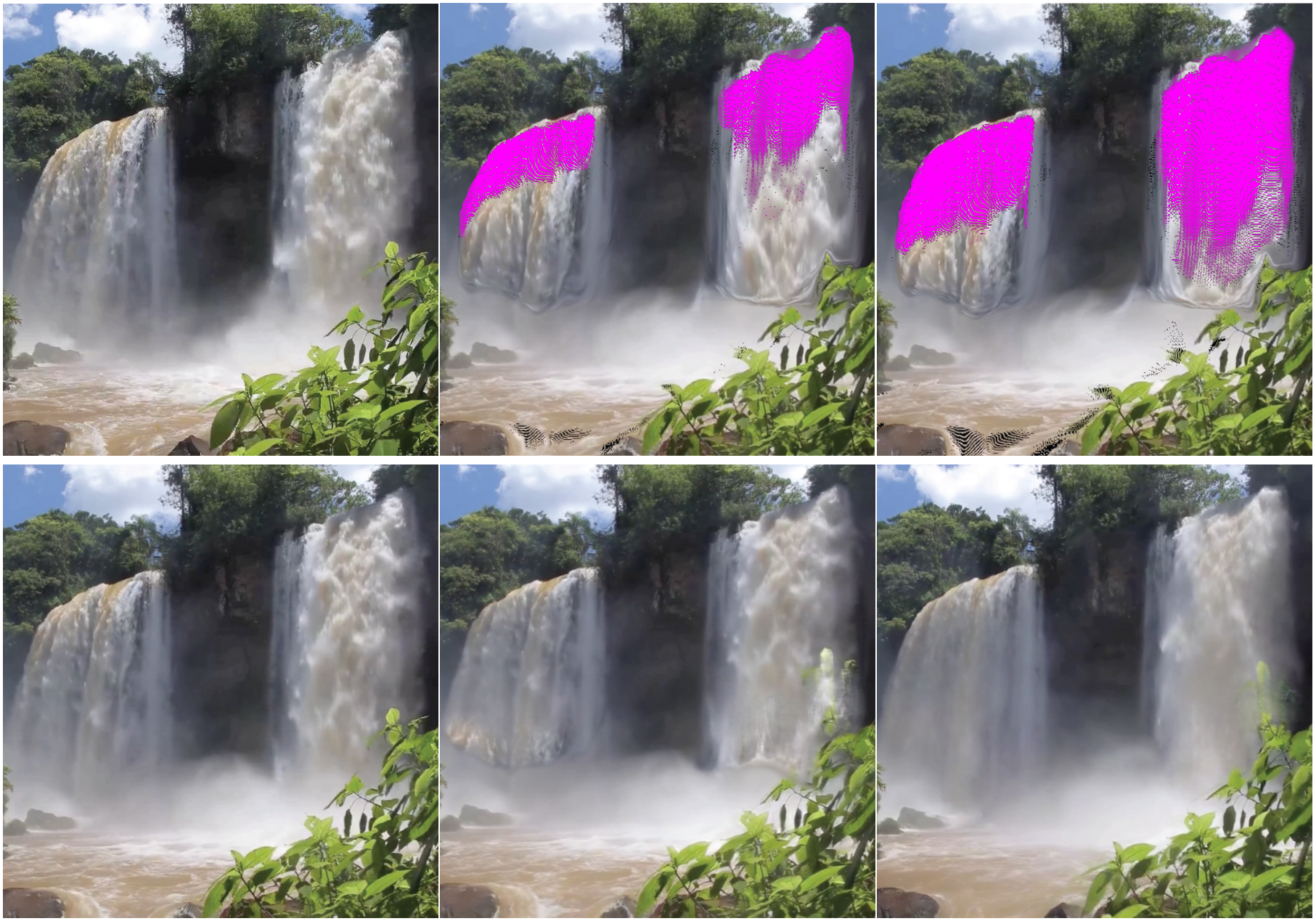}
    \caption{\small \textbf{Deep warping:} Above: Na\"ive splatting of RGB pixels results in increasingly large unknown regions over time, shown in magenta. Below: For the same frames, our deep warping approach synthesizes realistic texture in these unknown regions.}
    \label{fig:deep_warping}
\end{figure}
As feature pixels are warped through repeated integration of our motion field $M$, we typically observe increasingly large unknown regions (Figure~\ref{fig:deep_warping}), occurring when pixels vacate their original locations and are not replaced by others. This effect is especially prominent at motion ``sources", such as the top of a waterfall, where all predicted motion is outgoing. Although our decoder network is intended to fill these holes, 
it is still desirable to limit the complexity of the spatio-temporal inpainting task, as asking the network to animate an entire waterfall from a small set of distant features is unlikely to produce a compelling and temporally stable video. 

Our solution to this problem leverages the fact that our motion is \emph{textural} and \emph{fluid}, and thus 
much of the missing textural information in unknown regions can be feasibly borrowed from other parts of the frame that lie along the same motion path. With this intuition in mind, we describe a \emph{symmetric} splatting technique which uses reversed motion to provide valid textural information for regions which would otherwise be unknown. 

So far, the process we have described to generate an animated video involves warping the encoded feature map $D_0$ by $F_{0\rightarrow t}$ to produce future feature maps $V_f=\{D_0 ... D_N\}$, which are decoded to produce the output video frames.  However, since our motion map $M$ defines the motion between adjacent frames, we could just as easily animate the image by generating a video of the past, i.e., instead of warping $D_0$ into the future, use $-M$ to compute $F_{0\rightarrow -t}$, resulting in warped feature maps $V_p=\{D_{-N} ... D_0\}$. Decoding this feature video produces an equally plausible animation of the frame, with the main difference being that the large unknown regions in $V_p$ occur at the start of the sequence, as opposed to at the end of the sequence in $V_f$.

In fact, because the direction of motion has been reversed, the motion sources have been replaced with motion ``sinks" and vice versa (Figure~\ref{fig:crossfading}). This means that the locations of the unknown regions in $V_p$ are also largely complementary to those found in $V_f$. For instance, if our input image contains a waterfall, $V_f$ will begin with the input feature map $D_0$, and pixels will gradually flow down the waterfall, eventually accumulating at the bottom, and leaving a large unoccupied region at the top. Conversely, $V_p$ will begin with pixels accumulated at the top of the waterfall, and a large hole at the bottom, and will end with $D_0$. We leverage this complementarity by compositing pairs of feature maps (one in the past, one in the future) to produce a feature map which is typically fully dense. 

We perform this composition through joint splatting: we splat each pixel of $D_0$ twice to the same destination frame, once using $F_{0\rightarrow t}$ and once using $F_{0\rightarrow t-N}$. Note that  $F_{0\rightarrow t}$ does not necessarily equal $-F_{0\rightarrow -t}$, rather $F_{0\rightarrow -t}$ is the result of applying $-M$ recursively through Eq.~\ref{eq:eulerian}. As before, we use the softmax splatting approach with a network-predicted per-pixel weighting metric to resolve conflicts. This process results in a composite feature map that seldom contains significant holes, enabling generation of longer videos with larger magnitude motion.

\subsection{Looping} 
\label{sec:looping}
\begin{figure}
    \centering
    \includegraphics[width=\linewidth]{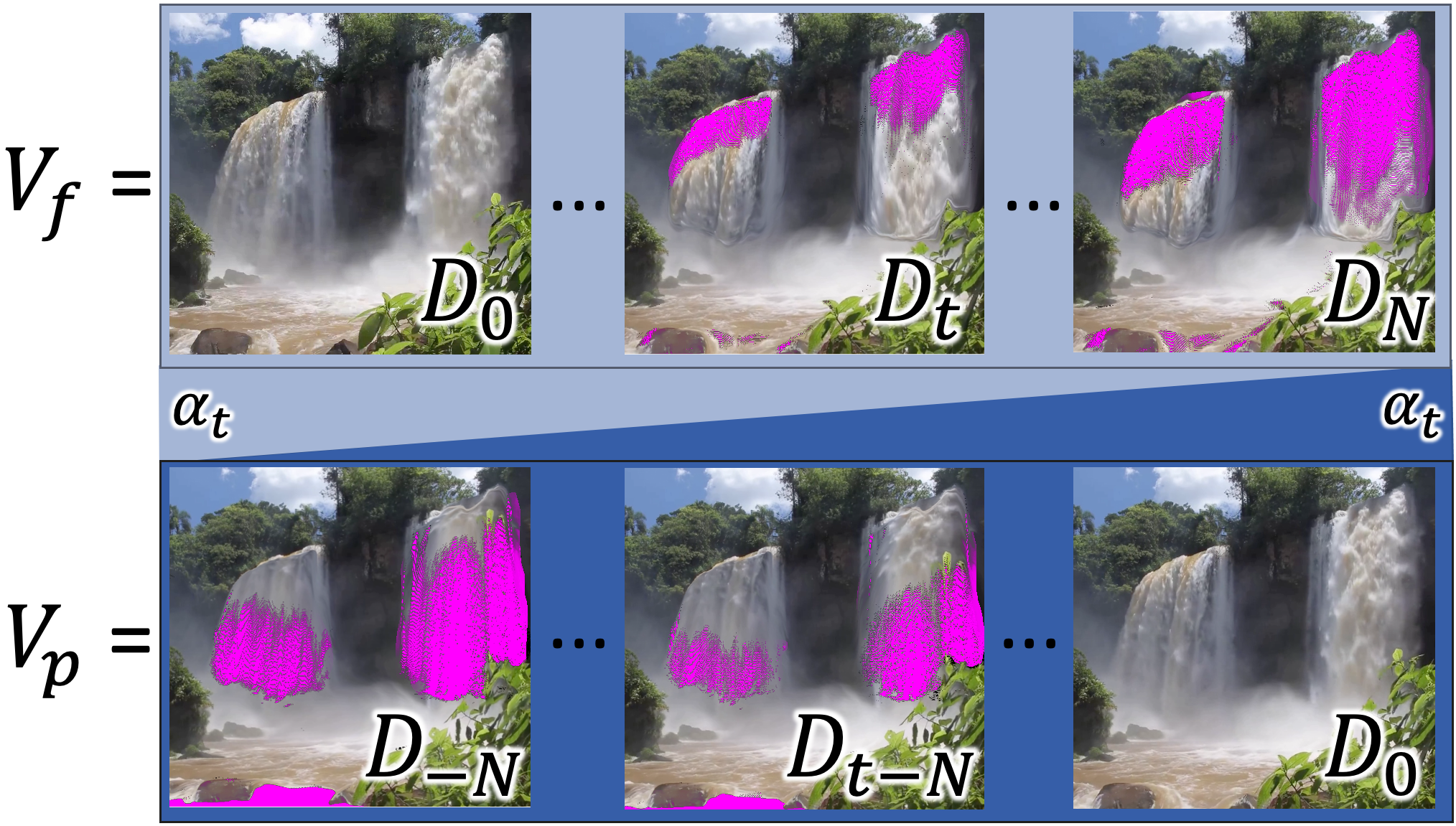}
    \caption{\small \textbf{Seamless looping}: An illustrated example of how seamless loops are created. Two feature videos are created by warping $D_0$.  The first, $V_f$, contains the result of integrating the motion field $M$, resulting in a video starting with the input image and animating into the future. The second, $V_p$, instead uses $-M$, resulting in a video starting in the past and ending with the input frame. These two videos typically contain complementary unknown regions (shown in magenta). Before decoding, we combine the two feature maps via joint splatting. We modulate the contribution of each using splatting weights $\alpha_t$, such that in the blended composite, the first and last frames are guaranteed to equal $D_0$, thus ensuring a seamless loop.
    Note that RGB images are shown for visualization, but these are both deep feature videos.}
    \label{fig:crossfading}
\end{figure}

In this section, we focus on ensuring that our output videos loop seamlessly. To this end, we first describe a modification to the splatting weights that guarantees that the first and last output video frames will be identical. Then, we describe an approach that enables end-to-end training without requiring a dataset of looping videos.

Prior work \cite{endo2019animatinglandscape} produces looping videos through crossfading: an animated, but non-looping, video is generated first, and a crossfade is applied across the two ends of the video to smooth out any jarring frame transitions. This approach can be quite effective in certain cases, but often produces artifacts in the form of double edges and ghosting. 

Instead of directly crossfading the animated video, our approach performs the transition in deep feature space, and provides the smoothly transitioning feature maps to the decoder. This allows us to enforce smooth transitions, while still producing images that contain realistic texture, avoiding many of the artifacts of direct crossfading. 

\paragraph{Looping weights.} Our looping technique relies on the observation that our two warped sequences $V_p$ and $V_f$ each have the input feature map $D_0$ on opposite ends of the sequence, as illustrated in Figure~\ref{fig:crossfading}. With this in mind, if we are able to smoothly control the contribution of each, such that the first frame contains only the values in $V_f$ and the last frame contains only the values in $V_p$, our feature maps (and our decoded images) on opposite ends of the video are guaranteed to be identical, and thus, our video is guaranteed to loop seamlessly. As such, we modulate the contribution of each feature map by introducing a temporal scaling coefficient to Eq.~(\ref{eq:splatting_metric}):
\begin{equation}
\label{eq:looping_metric}
D_t(\hat{\mathbf{x}}') = \frac{\sum_{\hat{\mathbf{x}}\in \mathcal{X}}  \alpha_t(\hat{\mathbf{x}}) \cdot D_0(\hat{\mathbf{x}}) \cdot \text{exp}(Z(\hat{\mathbf{x}}))}{\sum_{\hat{\mathbf{x}}\in \mathcal{X}} \alpha_t(\hat{\mathbf{x}}) \cdot \text{exp}(Z(\hat{\mathbf{x}}))}
\end{equation}
where $\mathcal{X}$ is the set of pixels which map to destination pixel $\hat{\mathbf{x}}'$, either by warping forward or backward in time. For a given frame $t$, we set:
\begin{equation}
\label{eq:alpha}
\alpha_t(\hat{\mathbf{x}}) =  \begin{cases} 
     \frac{t}{N} & \; \hat{\mathbf{x}} \in V_p \\
     1 - \frac{t}{N} & \;  \hat{\mathbf{x}} \in V_f
   \end{cases}
\end{equation}
Although the scaling coefficient $\alpha_t$ is linearly interpolated, the resulting composited feature video is not a linear interpolation of $V_p$ and $V_f$, since coinciding splatted features from each are typically not from the same input locations, and thus have different values of $Z$. Since the value of $Z$ is unconstrained and exponentiated, the overall magnitude of our weighting function ($\alpha_t(\hat{\mathbf{x}}) \cdot \text{exp}(Z(\hat{\mathbf{x}}))$) can vary significantly, and thus our composited feature map seldom contains equally blended features. The added coefficient $\alpha_t$ serves as a forcing function to ensure that the composited feature maps $D_t$ are equal to $D_0$ at $t=0$ and $t=N$, but composited features will transition from $V_f$ to $V_p$ at different rates per-pixel, depending on the relative magnitudes of the splatted $Z$ values.

\begin{figure}
    \centering
    \includegraphics[width=\linewidth]{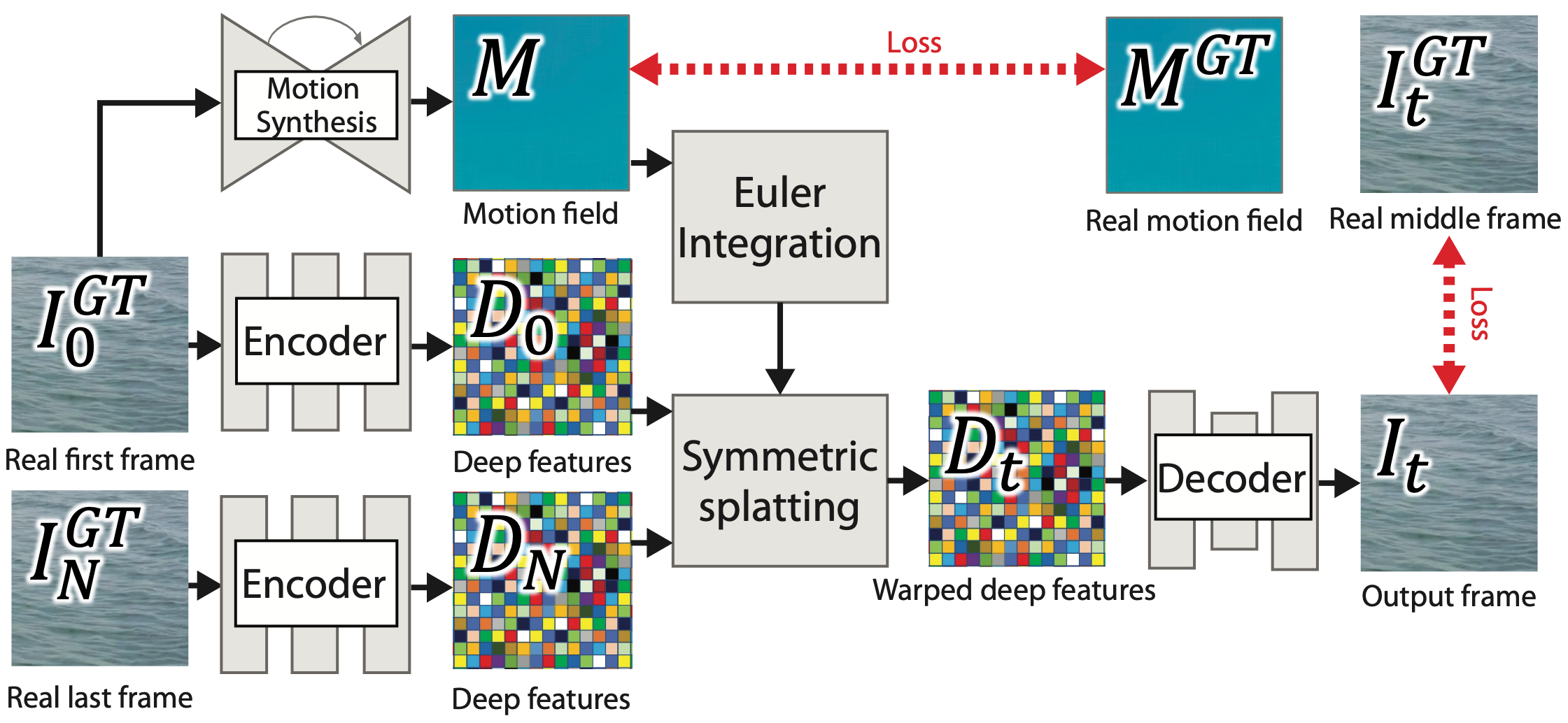}
\caption{\small \textbf{Training:} As described in Section~\ref{sec:warping}, each frame in our generated looping video is composed of textures from two warped frames. To supervise this process during training, i.e., to have a real frame to compare against, we perform our symmetric splatting using the features from two different frames, $I_0$ and $I_N$ (instead of $I_0$ twice, as in inference). We enforce the motion field $M$ to match the motion estimated from the ground truth video $M^{GT}$, and the output frame $I_t$ to match the real video frame $I^{GT}_t$. For both, we use a combination of photometric and discriminative losses.}
    \label{fig:training_pipeline}
\end{figure}

\paragraph{Training on regular videos.} Training our deep warping component (i.e., our encoder and decoder networks) to produce looping videos introduces an additional challenge: our training dataset consists of natural \emph{non-looping} videos. 
In other words, the looping video we are tasking our networks with generating does not exist, even for our training examples, and thus, it's non-trivial to formulate a reconstruction loss for supervision. 
Therefore, as illustrated in Figure~\ref{fig:training_pipeline}, we modify the task for training: instead of warping one frame in two directions, we use two different frames, one from the start of the video clip $I_0^{GT}$, and one from the end $I_N^{GT}$, encoded separately as feature maps. We additionally predict a motion field $M$ from $I_0^{GT}$, which is integrated to produce displacement fields $F_{0\rightarrow t}$ and $F_{0\rightarrow t-N}$. The two feature maps, $D_0$ and $D_N$, are respectively warped by $F_{0\rightarrow t}$ and $F_{0\rightarrow t-N}$ to an intermediate frame $t$, using our joint splatting technique with the weights defined in Eq.~\ref{eq:looping_metric}. Finally, the composited feature map $D_t$ is decoded to an image $I_t$, and a loss is computed against the real intermediate frame $I_t^{GT}$. At testing time, we perform the same process, except that instead of two input images, we use only one image, warped in both directions. This process is effectively training the network to perform video interpolation, and at inference time, using the network to interpolate between a frame and itself, while strictly enforcing the desired motion by warping the feature maps.

\section{Implementation Details}
\label{sec:implementation_details}
In this section, we provide more details about the implementation of our method. First, we provide a summary of the network architectures used for the motion estimation and warping networks. Then, we provide details about our training and inference pipelines. In order to facilitate future work, full training and inference code will be made publicly available at \href{eulerian.cs.washington.edu}{\texttt{eulerian.cs.washington.edu}}.

\paragraph{Network architecture.}
For the feature encoder and decoder networks, we use the architectures proposed in SynSin \cite{wiles2020synsin}, which have shown compelling results for single-image novel-view synthesis. Since our aim is not to generate new viewpoints, but rather to animate the scene, we replace the reprojection component with the softmax splatting technique proposed in Niklaus et al.~\cite{niklaus2020softmax}. Additionally, we replace the noise-injected batch normalization layer from SynSin with the modulated convolution approach proposed in Karras et al.~\cite{karras2020analyzing} (to which we also provide a latent noise vector). This modification greatly helps reduce visual artifacts and enables stable discriminator training with smaller batch sizes (a necessity for limited GPU memory).  For our motion estimation network, we use the architecture proposed in Pix2PixHD~\cite{wang2018high}.

\paragraph{Training.}
\begin{figure*}
    \centering
    \includegraphics[width=0.24\linewidth]{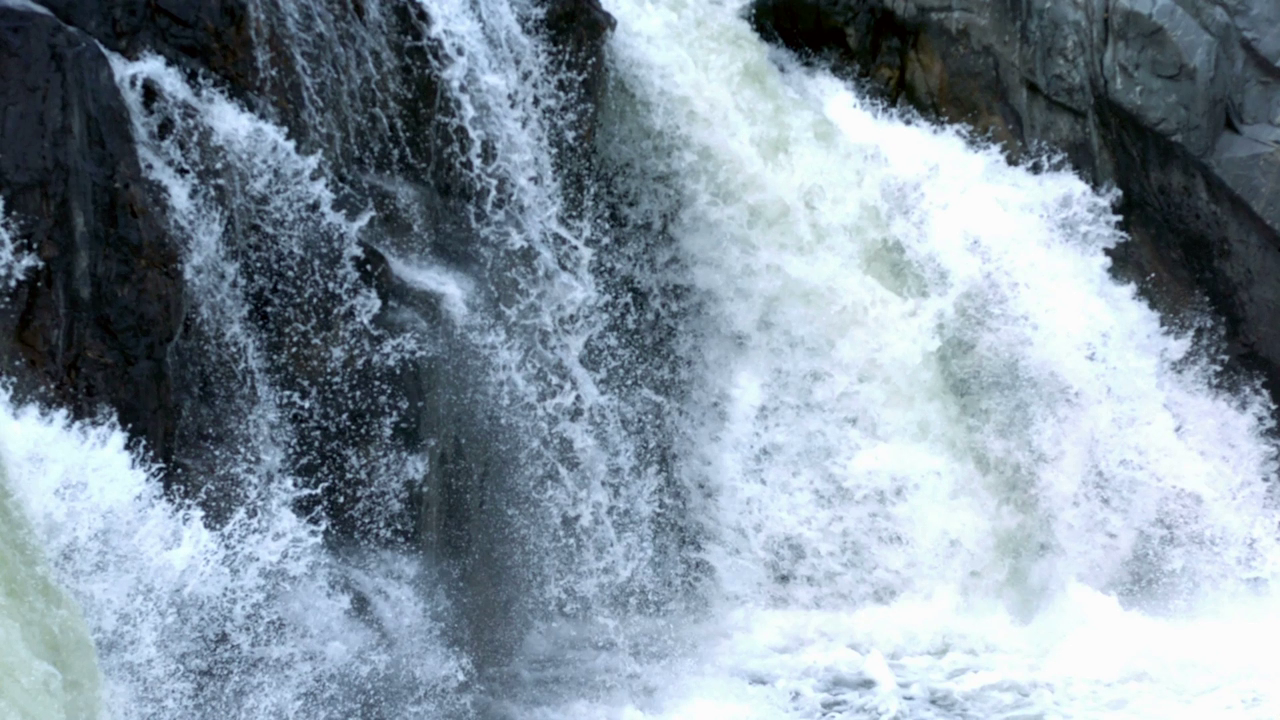} 
    \includegraphics[width=0.24\linewidth]{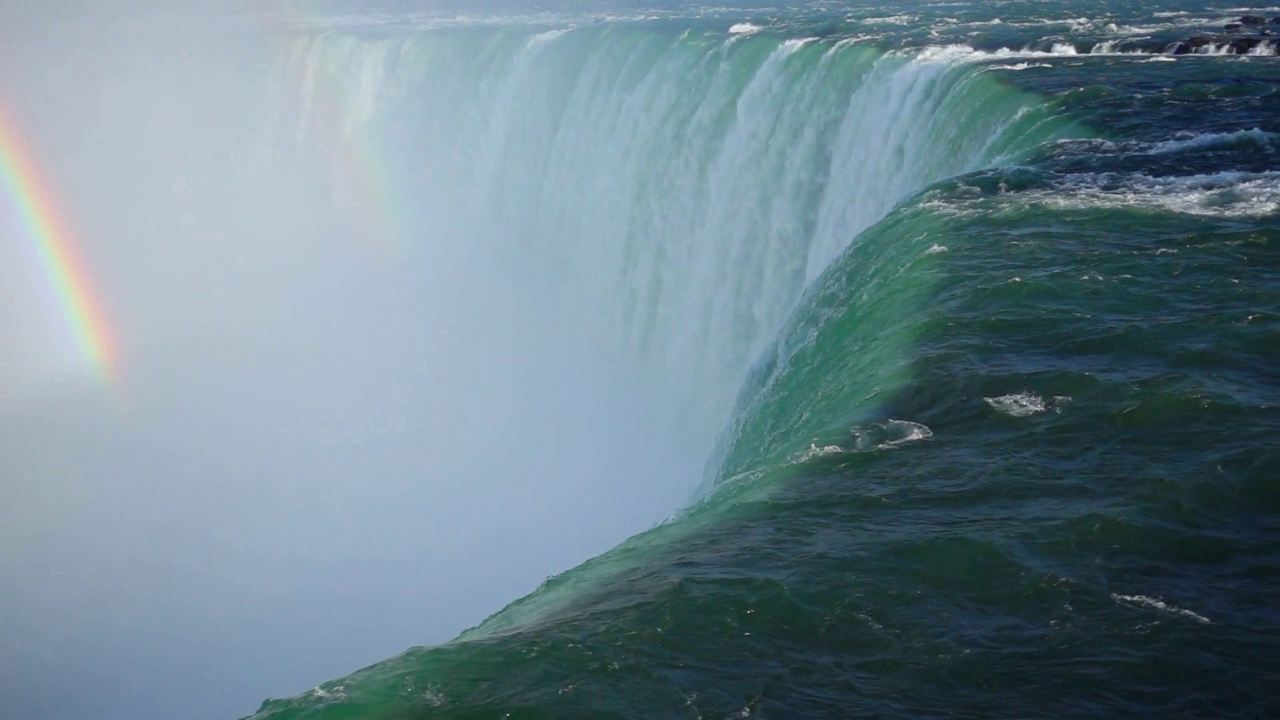} 
    \includegraphics[width=0.24\linewidth]{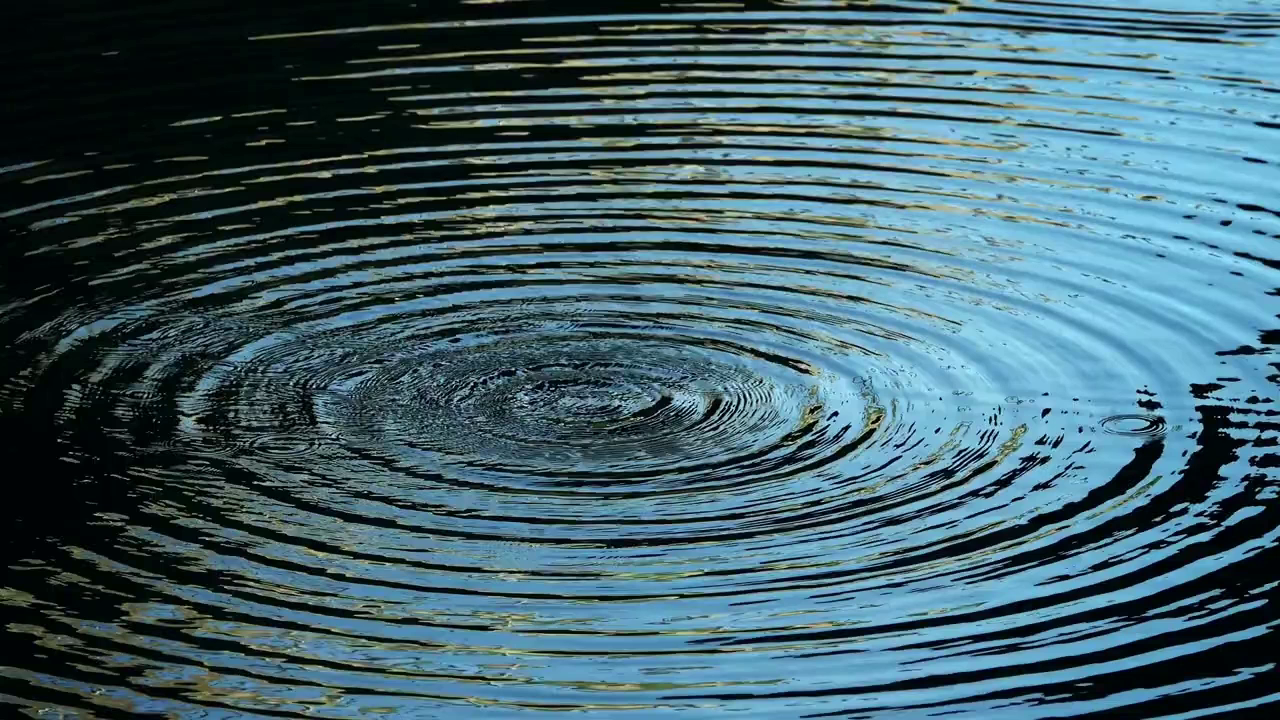} 
    \includegraphics[width=0.24\linewidth]{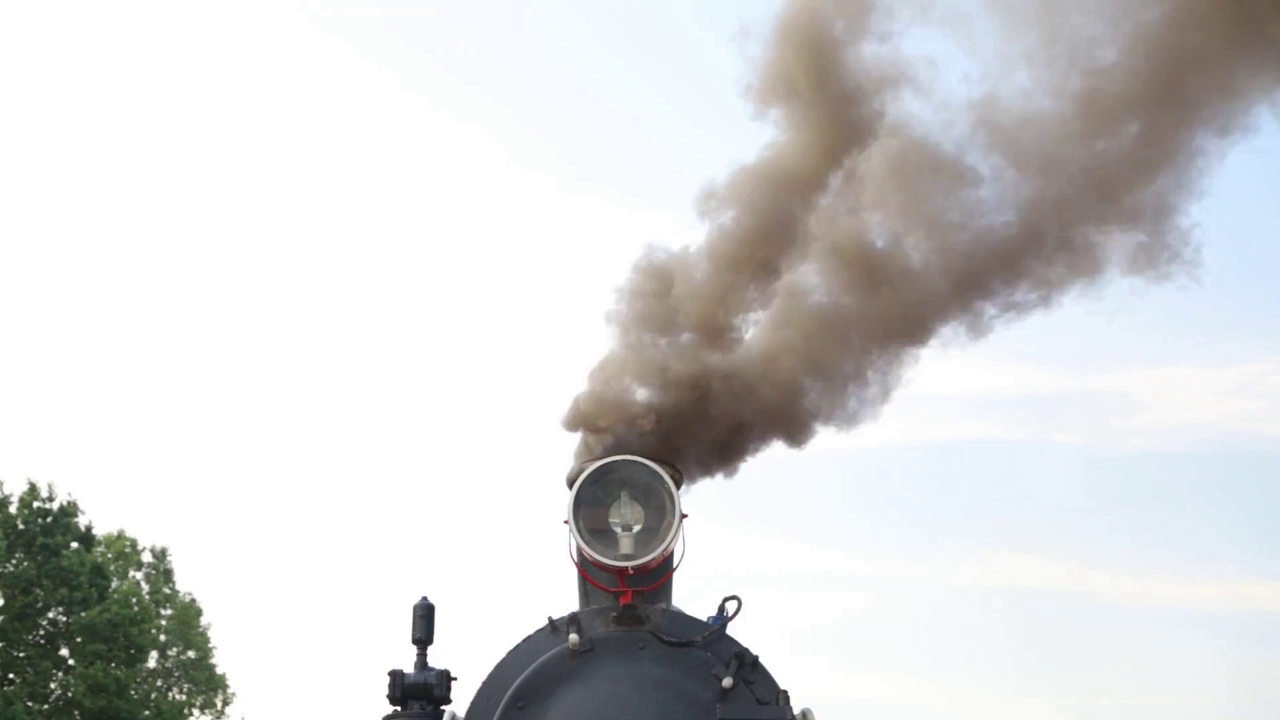}  \\
    \includegraphics[width=0.24\linewidth]{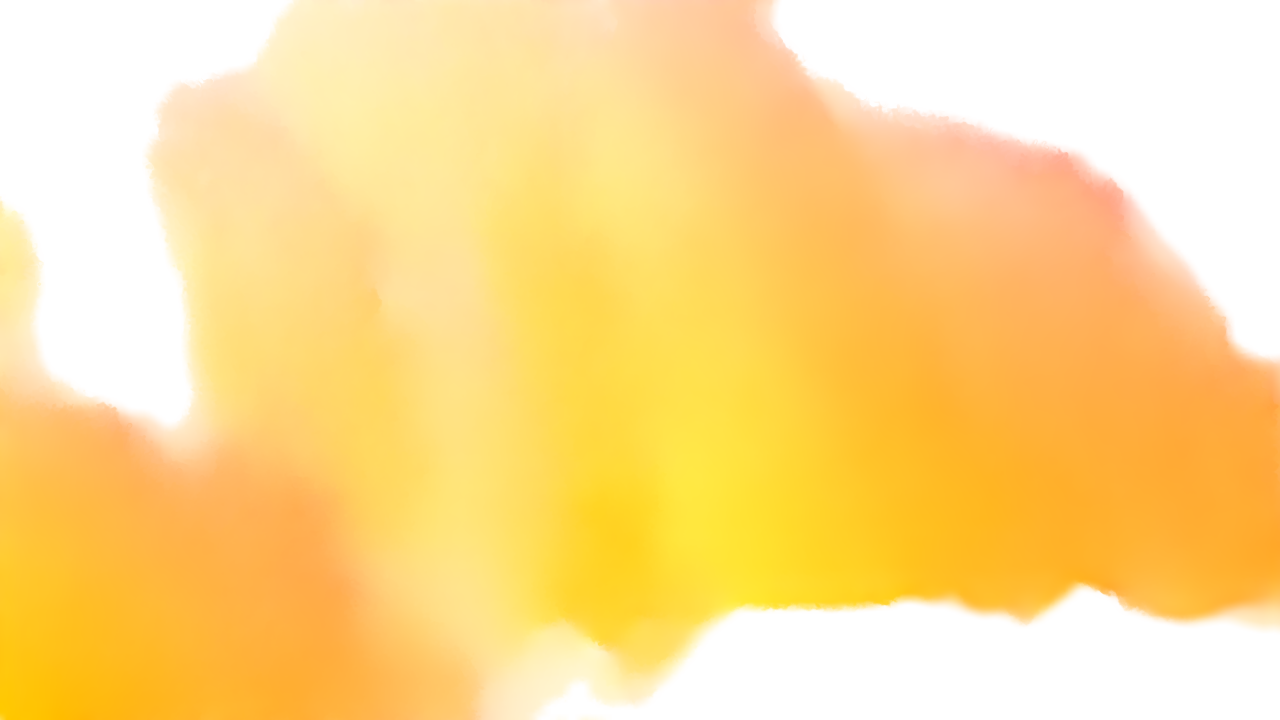} 
    \includegraphics[width=0.24\linewidth]{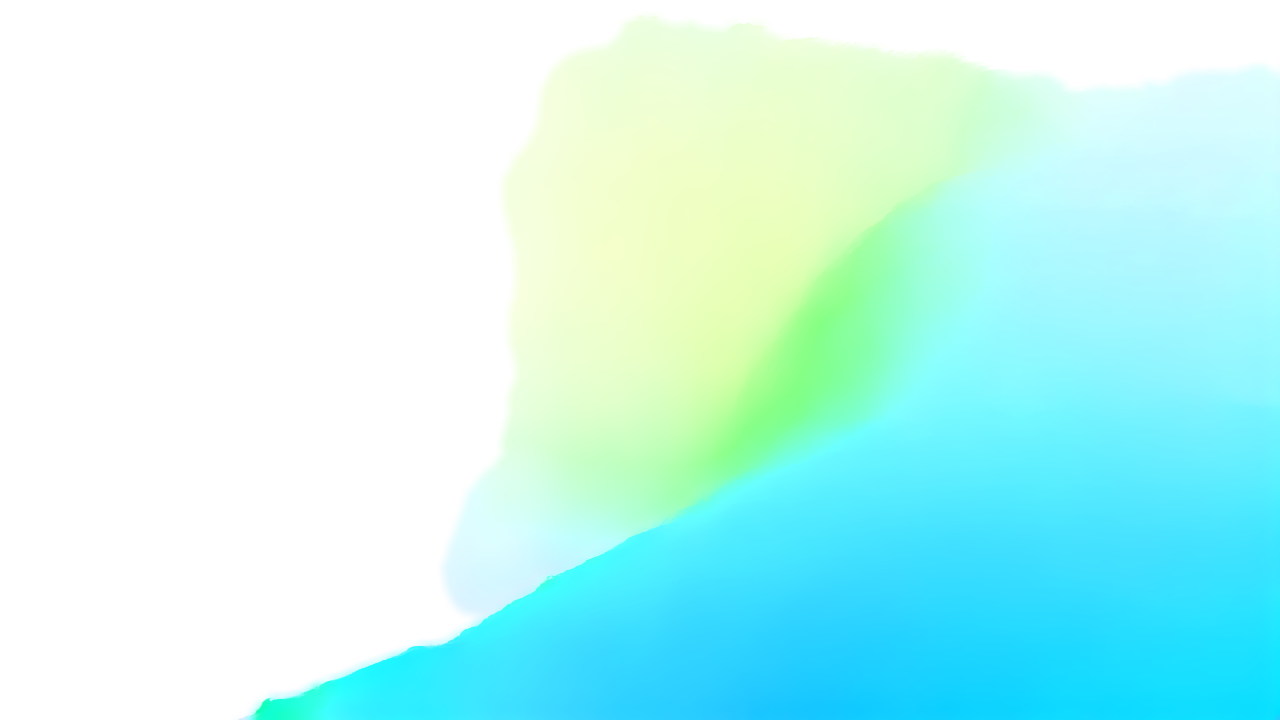} 
    \includegraphics[width=0.24\linewidth]{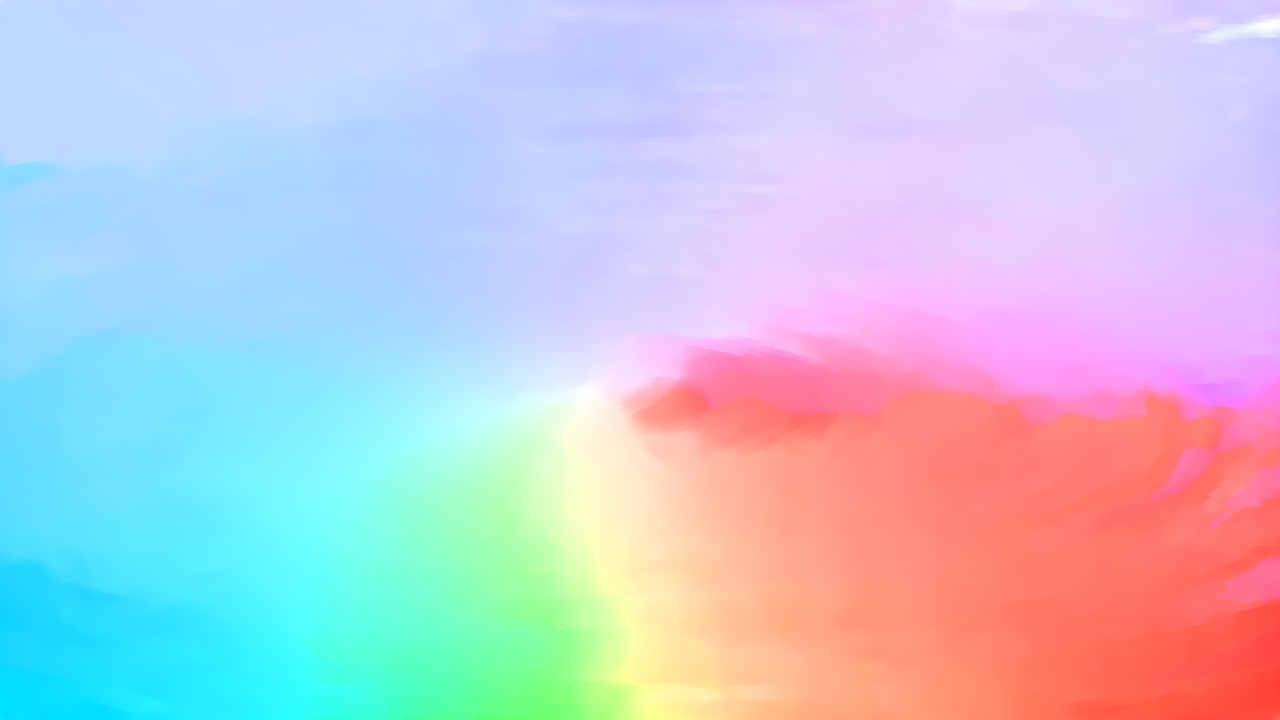} 
    \includegraphics[width=0.24\linewidth]{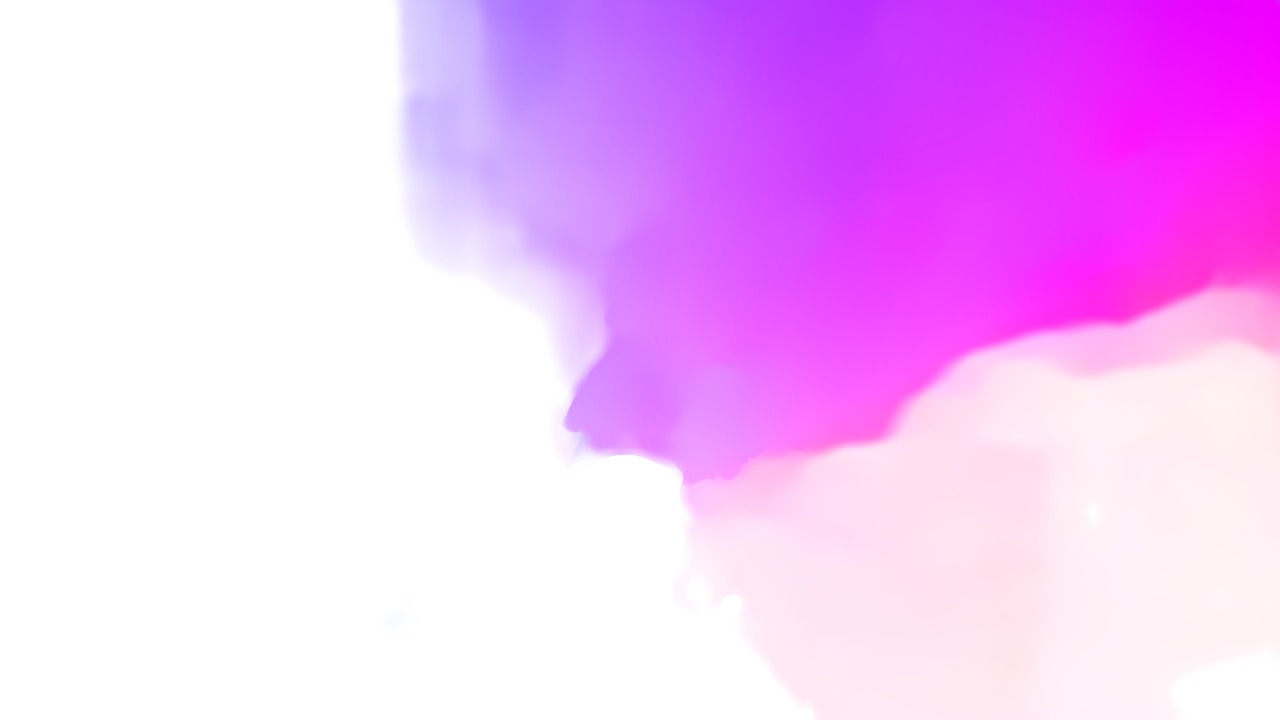}  \\
    \caption[Caption for LOF]{\small Examples of the input images (top), alongside their corresponding synthesized motion fields (bottom). Full resolution images, along with their corresponding animated videos, can be found in the supplementary video.}
    \label{fig:sample_data}
\end{figure*}

We focus on natural scenes with fluid textures such as waterfalls, turbulent streams, and flowing waves. For our training data, we collected and processed a set of 1196 unique videos of textural motion from an online stock footage website\footnote{www.storyblocks.com}. We use 1096 for training, 50 for validation, and 50 for testing. To generate ground-truth motion fields, we use a pre-trained optical flow estimator \cite{sun2018pwc} to compute the average optical flow between adjacent frames over a 2-second window. This effectively filters most motion which is cyclic, since pixels with cyclic motion will usually have been observed moving in opposing directions. 

We use only training videos from stationary cameras.

The motion estimation network is trained using 5 image-motion pairs from each of our 1096 training videos (a total of 5480 pairs) for 35 epochs, using the default parameters from Pix2PixHD~\cite{wang2018high}. Prior to training, we resize all our images to a standard size of $1280 \times 720$.  

The warping component is trained on 5 short video clips from each of our 1096 training videos. A training triplet (start frame, middle frame, end frame) is selected from each video clip at random during training, further increasing the effective dataset size. We also apply random augmentation to our training examples, including horizontal flips and cropping. We train the network on batches of 8 images of size $256\times 256$ for 200 epochs, using a discriminator learning rate of $3.5\times10^{-3}$ and generator learning rate of $3.5\times10^{-5}$. We use the same losses and loss balancing coefficients shown in SynSin \cite{wiles2020synsin}.

As in Niklaus et al.~\cite{niklaus2020softmax}, and for the purpose of training stability, we first train our two components separately. We start by training our motion estimation network supervised only by the ground-truth motion. Then, we train our warping component, which consists of the encoder and decoder networks, using the ground-truth motion fields as input. Finally, we fine-tune the two end-to-end, by warping with the predicted motion from the motion estimation network. This final step allows each network to best adapt to the properties and error characteristics of the other. We fine-tune for 20 epochs with discriminator and generator learning rates of $1\times10^{-3}$ and $1\times10^{-5}$.

\paragraph{Inference.} Our looping output videos have length $N=200$ with a framerate of $30$ frames per second. Each sequence is processed in 40 seconds on a Titan Xp GPU.

\section{Results \& Evaluation}

We first present a quantitative analysis of our method, and show comparisons with the state-of-the-art in still-image animation \cite{endo2019animatinglandscape} (Section~\ref{sec:quantitative}), as well as ablated variations of our method. Then, we show qualitative results of our method on a diverse collection of input images (Section~\ref{sec:qualitative}). We refer readers to our supplementary video for a full collection of visual results.

\subsection{Quantitative evaluation}
\label{sec:quantitative}

In this section, we present our experiments evaluating the different components of our method, i.e., (1) a novel motion representation, (2) a novel symmetric splatting technique, and (3) a novel looping technique.

\begin{figure}
    \centering
    \includegraphics[width=\linewidth]{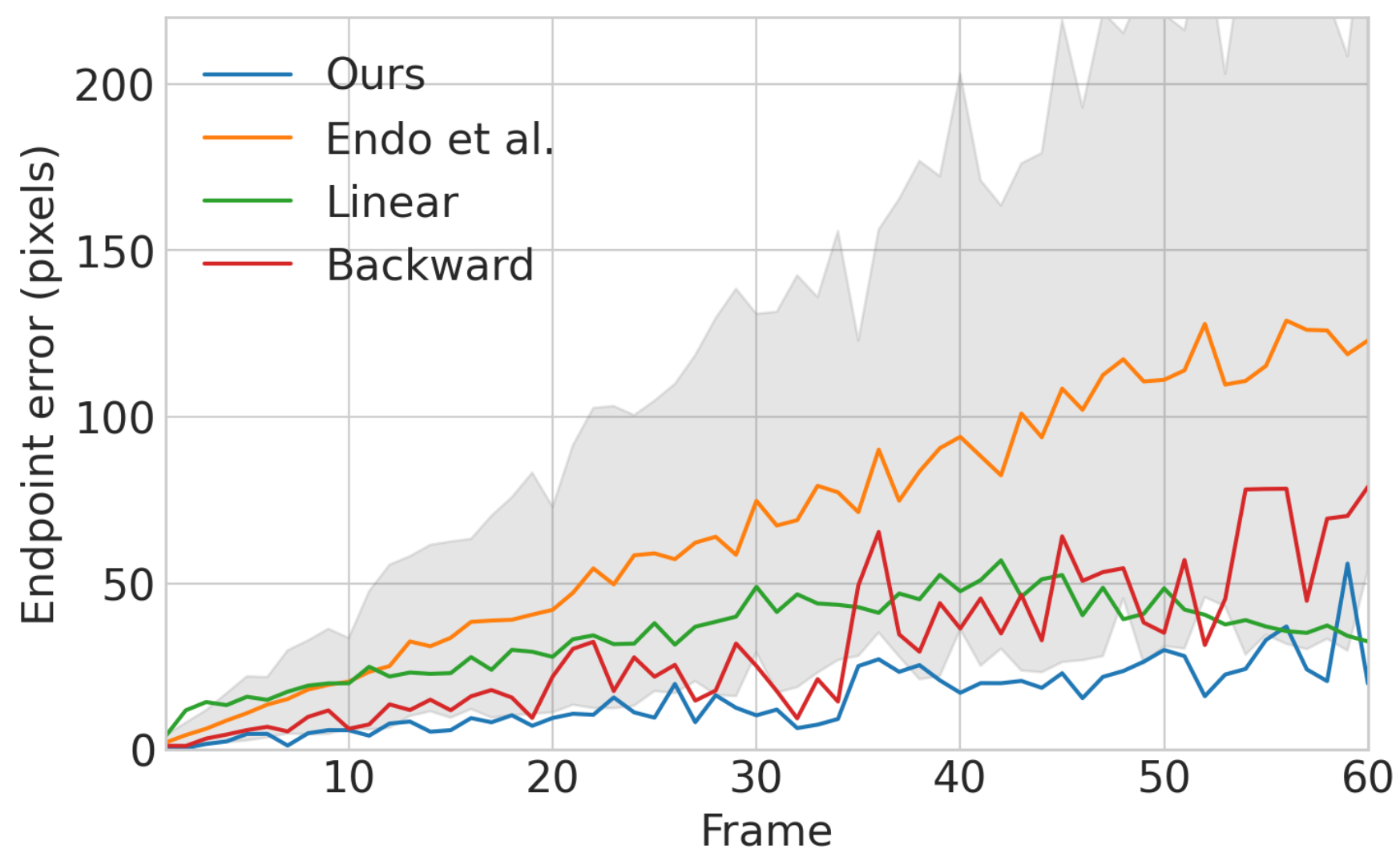}

    \caption{\small \textbf{Quantitative evaluation, motion prediction:} We evaluate the quality of the predicted motion by comparing the pixel positions in 60 future frames to those in the ground-truth video. We compare our proposed motion representation to three alternative methods, described in Section~\ref{sec:quantitative}. The shaded region shows the range of predictions produced by Endo et al.~\cite{endo2019animatinglandscape}. We find that our proposed motion representation is able to most reliably reproduce the true motion information for scenes with fluid motion. All comparisons are performed on images of size $1280\times720$. }
    \label{fig:motion_error}
\end{figure}

\paragraph{Motion representation.}
We evaluate the effectiveness of our proposed motion representation (integrated Eulerian flow) by comparing our predicted motion to ground truth pixel positions in future frames of the video. We establish ground truth motion by densely tracking all pixels through a sequence of 60 frames, using an off-the-shelf optical flow estimator \cite{sun2018pwc}. We report the average Euclidean error between the ground truth positions and those estimated through our synthesized motion field, i.e., the endpoint error. We compare our proposed method to the following variants: (1) the per-frame recurrent estimation from Endo et al.~\cite{endo2019animatinglandscape}, (2) directly predicting $F_{0\rightarrow N}$ and linearly interpolating intermediate motion $F_{0\rightarrow t}$ as $\frac{t}{N}F_{0\rightarrow N}$, and (3) training our motion network to predict the backward flow field, i.e., $M = F_{1\rightarrow 0}$ (and thus all splatting is replaced by backward warping).  The results of this experiment can be found in Figure~\ref{fig:motion_error}. We see that our method is able to most faithfully reproduce the ground-truth motion for our scenes. Empirically, we observe that the methods employing backward warping produce a majority of errors at motion boundaries, such as occlusions. We hypothesize that these differences are because the network is more easily able to predict an output that is spatially aligned with the input image. 

Since images may often have many plausible motion directions, we ensure that the comparisons performed in this experiment are on video examples that contain unambiguous motion, eg. waterfalls and rivers. In order to identify these samples, we gave each of 5 users 50 images from different scenes, and asked them to manually annotate the likely motion direction. We retained the scenes for which all the annotations were within 30 degrees of the median motion direction in our ground truth flow values. Additionally, since we prefer the motion comparison to be agnostic to animation speed, i.e., animating the scene realistically but in slow-motion is acceptable, we solve for a per-sequence time-scaling constant that best aligns the motion magnitudes of the predicted and ground-truth displacement fields. This constant is computed for all methods and is used in all our comparisons. 

The motion estimation network from Endo et al.~\cite{endo2019animatinglandscape} uses a latent code as input to the network, and different latent codes produce different predicted motions. 
To consider all possible outcomes of their method, we randomly sample 100 latent codes from the training codebook and report statistics on the resulting synthesized motions.

\paragraph{Video synthesis.}

\begin{table}
    \centering
    \scalebox{0.80}{
\begin{tabular}{l|l|l|lll}
\toprule
                       & $\uparrow$ PSNR  & $\uparrow$  SSIM &  $\downarrow$ LPIPS &  &  \\ \hline
Na\"ive color splatting  &   7.90  &   0.313   &   0.595    &  &  \\ \hline
Backward Warping        &   10.29   &  0.409    &  0.483     &  &  \\ \hline
Ours - $Z(\hat{\mathbf{x}}) = 1$  &   13.88   &  0.541  &  0.344     &  &  \\ \hline
Ours - No Symmetric Splatting &  12.19    &   0.493   & 0.418 &  &  \\ \hline
\textbf{Ours - Full}                   &  \textbf{14.63}    &   \textbf{0.619}   &  \textbf{0.313}     &  &  \\
\bottomrule
\end{tabular}}

    \caption{\small \textbf{Quantitative evaluation, video synthesis}: We evaluate the quality of future frame predictions by comparing 60 synthesized frames with corresponding frames in the ground truth video. We compare our method to four alternatives, described in Section~\ref{sec:quantitative}. All variant use our proposed motion estimation network.}
    \label{fig:synthesis_error}
\end{table}
Second, we evaluate the choice of warping technique. Given the same flow values for a set of testing (unseen) video clips, we evaluate five future frame synthesis techniques: (1) na\"ive color splatting, where the feature encoder and decoder are not used, and instead the color values are warped, (2) backward warping, where the forward displacement field is inverted \cite{shade1998layered}, and then backward warping is applied, such that no holes occur during warping, (3) our method without the network inferred weights, i.e., $Z(\hat{\mathbf{x}})=1$ for all pixels, (4) our method without symmetric splatting, and (5) our full method. We again use temporally-scaled sequences with unambiguous motion to compare each method's synthesized frames with the ground truth future video frames. We perform this comparison using PSNR, SSIM, and LPIPS~\cite{zhang2018unreasonable}. Table~\ref{fig:synthesis_error} shows a quantitative comparison of these techniques, demonstrating that our proposed approach outperforms the alternatives at synthesizing future frames when the same motion is provided. Additionally, in the supplemental material, we show a qualitative comparison of these techniques. Compared to our approach, we observe that standard color splatting results in significant sparsity, i.e. many holes with unknown color. Backward warping instead fills these holes with interpolated (stretched) texture, which in most cases is equally jarring. Feature warping without inferred $Z(\hat{\mathbf{x}})$ values results in blurred details, since features are more often evenly combined. Removing symmetric splatting results in large unknown regions, which are filled in by the decoder network with blurry and often unrealistic texture.

\paragraph{Looping.}
\begin{table}
\centering
\scalebox{0.85}{
\begin{tabular}{llllll}
\toprule
                   & $S\geq 1$ & $S\geq 2$ & $S = 3$  \\ \hline
\multicolumn{1}{l|}{Endo et al. \cite{endo2019animatinglandscape}}  & \multicolumn{1}{l|}{348} & \multicolumn{1}{l|}{101} & \multicolumn{1}{l}{9} &  \\ \hline
\multicolumn{1}{l|}{Ours - No $\alpha_t$}          & \multicolumn{1}{l|}{173} & \multicolumn{1}{l|}{183} & \multicolumn{1}{l}{0} &  \\ \hline
\multicolumn{1}{l|}{Ours - Crossfade}               & \multicolumn{1}{l|}{470} & \multicolumn{1}{l|}{418} & \multicolumn{1}{l}{43} &  \\ \hline
\multicolumn{1}{l|}{\textbf{Ours - Full}} & \multicolumn{1}{l|}{\textbf{500}} & \multicolumn{1}{l|}{\textbf{472}} & \multicolumn{1}{l}{\textbf{448}} &  \\ 
\bottomrule
\end{tabular}
}
    \caption{\small \textbf{User study}: We perform a user study to compare four techniques for producing looping videos. We collected 5 unique annotations for each of 100 samples. We direct users to judge the visual quality and realism of each looping video and rank the videos with unique scores $S=[0,3]$, where $3$ is best. We report the cumulative number of annotations above a certain ranking. On average, users rank our method higher than the alternatives. }
    \label{fig:user_study}
\end{table}
Finally, we evaluate the choice of our looping technique. We compare four approaches: (1) our synthesis technique followed by the crossfading technique described in Endo et al.~\cite{endo2019animatinglandscape}, (2) the end-to-end pipeline described in Endo et al.~\cite{endo2019animatinglandscape}, (3) our approach without the scaling coefficient $\alpha_t$ introduced in Eq.~\ref{eq:looping_metric} and (4) our proposed approach. Since we do not have a ground truth looping video for comparison, we instead perform a user study, in which 100 MTurk users are asked to rank the four variants by visual quality. Table~\ref{fig:user_study} shows the results of the user study, which demonstrate that our proposed approach compares favorably against the alternatives. In the supplemental material, we also show a visual comparison of these approaches. For our comparison to Endo et al.~\cite{endo2019animatinglandscape}, we use an implementation provided by the authors, trained on our dataset for the recommended 5000 epochs. Note that we only use the motion estimation component of their method, as our scenes are not timelapses, and therefore do not have as significant changes in overall appearance.

\subsection{Qualitative evaluation}
\label{sec:qualitative}

For evaluation purposes, we demonstrate our system on a large collection of still images. A subset of these images, along with their synthesized motions, can be seen in Figure \ref{fig:sample_data}. The dataset contains a variety of natural scenes, including waterfalls, oceans, beaches, rivers, smoke, and clouds.
In the supplementary video, we provide a larger set of input images and final rendered animations, as well as intermediate outputs such as synthesized motion fields.

In the results, we can see that the network learns important motion cues, such as perspective (i.e. motion is larger for objects closer to the camera), water turbulence, and detailed flow direction from surface ripples. By comparison, we find that the generated videos using the method in Endo et al.~\cite{endo2019animatinglandscape} more often produces videos with unrealistic motion or incorrect motion boundaries. Additionally, since our method performs warping in the deep feature domain, instead of explicitly warping RGB pixels, our results do not contain many of the same characteristic artifacts of warping, such as shearing or rubber-sheeting. Finally, we observe that our results loop more seamlessly, without obvious crossfading or ghosting.

\section{Conclusion}

In this paper, we have presented a method that can synthesize realistic motion from single photographs to produce animated looping videos. 
We introduced a novel motion representation for single-image textural animation that uses Euler integration. This motion is used to animate the input image through a novel symmetric splatting technique, in which we combine texture from the future and past. Finally, we introduced a novel video looping technique for single-frame textural animation, allowing for seamless loops of our animated videos.

We demonstrated our method on a wide collection of images with fluid motion, and showed that our method is able to produce plausible motion and realistic animations.

{\small
\bibliographystyle{style/ieee_fullname}
\bibliography{references.bib}
}

\end{document}